\documentclass{article}

\usepackage{microtype}
\usepackage{graphicx}
\usepackage{subcaption}
\usepackage{booktabs} 
\usepackage{tabularx}

\usepackage{svg}
\usepackage{tikz}
\usetikzlibrary{positioning, arrows.meta, shapes.geometric, fit,
                backgrounds, calc, decorations.pathreplacing}
\usepackage{xcolor}
\usepackage{graphicx}
\graphicspath{{paper/figures/}}   

\usepackage{hyperref}

\usepackage{listings}
\lstdefinelanguage{json}{
    morestring=[b]",
    morecomment=[l]{//},
    commentstyle=\color{gray},
    keywords={null, true, false},
    keywordstyle=\color{blue}\bfseries,
    stringstyle=\color{red},
    literate=
        *{0}{{{\color{black}0}}}{1}
        {1}{{{\color{black}1}}}{1}
        {2}{{{\color{black}2}}}{1}
        {3}{{{\color{black}3}}}{1}
        {4}{{{\color{black}4}}}{1}
        {5}{{{\color{black}5}}}{1}
        {6}{{{\color{black}6}}}{1}
        {7}{{{\color{black}7}}}{1}
        {8}{{{\color{black}8}}}{1}
        {9}{{{\color{black}9}}}{1}
        {:}{{{\color{black!60}{:}}}}{1}        
        {\{}{{{\color{black!60}{\{}}}}{1}      
        {\}}{{{\color{black!60}{\}}}}}{1}      
        {[}{{{\color{black!60}{[}}}}{1}        
        {]}{{{\color{black!60}{]}}}}{1}        
        {,}{{{\color{black!60}{,}}}}{1},       
}
\usepackage{xcolor} 

\usepackage[preprint]{icml2026}

\usepackage{amsmath}
\usepackage{amssymb}
\usepackage{mathtools}
\usepackage{amsthm}
\usepackage{booktabs}

\usepackage[capitalize,noabbrev]{cleveref}

\theoremstyle{plain}

\theoremstyle{definition}

\theoremstyle{remark}

\usepackage[textsize=tiny]{todonotes}

\begin{document}

\twocolumn[
  \icmltitle{Bridging the Last Mile of Time Series Forecasting with LLM Agents}



  \icmlsetsymbol{equal}{*}

  \begin{icmlauthorlist}
    \icmlauthor{Yuhua Liao}{comp}
    \icmlauthor{Zetian Wang}{comp}
    \icmlauthor{Qiangqiang Nie}{comp}
    \icmlauthor{Zhenhua Zhang}{comp}
  \end{icmlauthorlist}

  \icmlaffiliation{comp}{Trip.com Group, Shanghai, China}

  \icmlcorrespondingauthor{Yuhua Liao}{yh\_liao@trip.com}

  \icmlkeywords{Machine Learning, ICML}

  \vskip 0.3in
]



\printAffiliationsAndNotice{}  

\begin{abstract}
Time series forecasting has advanced rapidly, especially with the emergence of foundation models that show strong zero-shot performance on numerical extrapolation. However, in real-world forecasting settings, a statistically plausible baseline is rarely the final forecast used in practice. Before a forecast becomes decision-ready, it often needs to be revised using weakly structured business context such as holiday effects, campaign plans, external events, historical analogs, and expert feedback. This practical stage remains underexplored in the forecasting literature. In this paper, we formulate this stage as the \textbf{last-mile forecasting} problem and present an LLM-agent framework that sits on top of a forecasting backbone. Our system maintains a unified forecast workspace, invokes tools to retrieve contextual evidence, and converts reasoning trajectories into explicit forecast revision actions under structural safety constraints. It also supports long-horizon forecasting through map-reduce-style decomposition and post-hoc reflection through a memory bank. The resulting system is designed to be controllable and auditable. Through real-world case studies, we show how LLM agents can bridge the gap between statistical prediction and business-ready forecasting.
\end{abstract}

\section{Introduction}

Time series forecasting has progressed rapidly in recent years. Foundation models and large pretrained sequence models have substantially improved the quality and zero-shot usability of forecasts across domains \cite{shi2026kronos,li2025mira}. These advances have made it increasingly practical to obtain strong numerical baselines from historical observations alone \cite{rasul2023lag,das2023decoder,ansari2024chronos,liang2024foundation}. However, in many operational settings, a statistically plausible baseline is not yet a decision-ready forecast. Before forecasts are used for planning or resource allocation, they are often revised to account for contextual factors that are absent from the observed series, such as holiday calendars, promotional schedules, external disruptions, policy changes, historical analogs, and expert judgment \cite{lawrence2006judgmental,fildes2009effective}.

Post-baseline forecast adjustment has long been studied in judgmental forecasting. However, it is usually treated as a human or organizational process, rather than as a structured revision problem that can be executed and audited by an agent. Standard forecasting research primarily improves the predictor that maps historical observations to future values. Multimodal forecasting broadens the predictive input space by incorporating text, metadata, or other auxiliary signals \cite{kim2024multi,jiang2025multi}. Recent LLM-based systems further show that agents can support retrieval, tool use, iterative refinement, and broader time-series reasoning \cite{citation-0,yeh2025empowering,jalori-etal-2025-flairr,zhao2025timeseriesscientist}. These directions can all improve forecast quality, but they typically focus on how forecasts are generated or refined, rather than on how an already-generated baseline should be contextually revised, justified, and tracked under operational constraints.

This post-baseline setting matters because a useful revision process must support more than numerical accuracy. A planner may need to know why a holiday uplift was applied, which evidence supported it, whether historical observations were preserved, whether the original baseline remains recoverable, and how an adjustment performed once actual values arrived \cite{fildes2009effective}. In such settings, the forecast is only one part of the output; the revision process itself must also be controllable and auditable. These requirements are difficult to satisfy when an LLM is asked to directly emit a replacement forecast as free-form output.

We refer to this setting as \textbf{last-mile forecasting}: the problem of transforming a statistically plausible baseline into a decision-ready forecast through context-aware revision. The key challenge is not to replace the forecasting backbone, but to build a reasoning framework that can interpret heterogeneous evidence and convert contextual judgments into explicit forecast-editing actions. Under this view, the output is not merely a future series. It is a pair consisting of a forecast and a revision trace that records how the forecast changed, why it changed, and which evidence supported the change.

To study this problem, we present an LLM-agent framework that sits on top of a forecasting backbone. Our contributions are threefold. First, we formulate last-mile forecasting as a distinct systems problem between baseline generation and business-ready forecast consumption. Second, we propose an action-centric agent framework that converts contextual reasoning into constrained actions on a shared forecast workspace. Third, through real-world case studies, we demonstrate that the proposed framework improves forecasting accuracy while enabling controllable, auditable, and iterative refinement.

\section{Related Work}

\subsection{Judgmental Forecasting and Forecast Adjustment}
Judgmental forecasting studies how human judgment and domain knowledge complement statistical forecasts, especially when future outcomes depend on contextual information outside the observed series, such as promotions, competitor activity, managerial knowledge, or special events \cite{lawrence2006judgmental,webby1996judgemental}. In operational planning, computerized systems often produce initial statistical forecasts that are later reviewed and adjusted by demand planners \cite{fildes2009effective}. Prior work has studied this integration through voluntary adjustment, mechanical combination, and forecasting support systems \cite{goodwin2002integrating,goodwin2007process,webby2005forecasting}, while also showing that manual adjustments can introduce bias, overreaction, or undocumented interventions \cite{fildes2009effective,goodwin2002integrating}. We build on this tradition but shift the computational setting: post-baseline revision is represented as constrained, evidence-backed actions executed by an LLM agent.

\subsection{Time Series Foundation Models}

Recent work has shown that foundation-model paradigm transfers effectively to time series analysis. Lag-Llama, TimesFM, Chronos, Moirai, and related surveys demonstrate the potential of pretrained models to provide one-fits-all backbones with strong zero-shot or few-shot performance across domains \cite{rasul2023lag,das2023decoder,ansari2024chronos,woo2024unified,liang2024foundation}. These models shift forecasting from domain-specific pipelines toward reusable forecasting capabilities. At the same time, recent position work has questioned whether a single architecture can fit all forecasting domains, arguing that domain-specific structure and operational context remain difficult to absorb into a universal forecasting model \cite{ma2026position}.

\subsection{LLMs and Agents for Time Series Reasoning}

LLMs have also been adapted directly to time series tasks. PromptCast frames forecasting as text generation, while Time-LLM and related approaches reprogram or prompt language models for numerical prediction \cite{jin2023time,xue2023promptcast}. Multimodal forecasting further incorporates textual or metadata signals into prediction pipelines \cite{kim2024multi,jiang2025multi}. These methods show that language models and auxiliary modalities can improve forecasting.

More recent work goes beyond direct prompting and casts LLMs as autonomous agents that plan, invoke tools, and iteratively refine their outputs \cite{yao2023react,shinn2023reflexion}; this agentic paradigm has been transferred to time series. Systems such as TS-Reasoner and recent agentic forecasting frameworks use tools, retrieval, and iterative refinement for time series analysis and prediction \cite{citation-0,yeh2025empowering,jalori-etal-2025-flairr,zhao2025timeseriesscientist}. Our setting is narrower and more operational: the LLM is not the numerical forecasting backbone, but a reasoning-and-orchestration system that maps contextual evidence to explicit, auditable forecast revision actions. This places our work between judgmental forecast adjustment and agentic time-series systems, with the specific focus on the last mile between baseline prediction and business-ready forecasting.

\section{Problem Formulation}
\label{sec:problem}

Conventional forecasting is typically formulated as direct sequence prediction. Given a historical series
\begin{equation}
    X_{1:T} = (x_1, x_2, \dots, x_T),
\end{equation}
the goal is to predict a future horizon of length \(H\):
\begin{equation}
    \hat{Y}_{T+1:T+H} = \mathcal{F}(X_{1:T}).
\end{equation}
In last-mile forecasting, the backbone forecast is not the final object of interest. We assume that a forecasting model first produces a baseline forecast
\begin{equation}
    F_{\text{base}} = \mathcal{F}(X_{1:T}) \in \mathbb{R}^{H},
\end{equation}

and this baseline must then be revised using contextual information $C = \{c_i\}_{i=1}^{m}$, where each \(c_i\) denotes information outside the time series itself, such as a user instruction, a calendar event, a retrieved historical analog, an external signal, or expert feedback. The objective is not to regenerate a forecast from scratch, but to transform \(F_{\text{base}}\) into a final forecast through an inspectable sequence of revisions.

We formulate this process as constrained sequential revision over a forecast workspace. At revision step \(t\), the workspace state is

\begin{equation}
    W_t = (X_{1:T}, F_{\text{base}}, F_t, E_t, A_t),
\end{equation}

where \(F_t \in \mathbb{R}^{H}\) is the current editable forecast, \(E_t\) stores structured evidence derived from the contextual inputs \(C\) and any tool observations, and \(A_t = (a_0,\dots,a_{t-1})\) is the accumulated action trace. The initial state satisfies \(F_0 = F_{\text{base}}\), \(E_0 = \emptyset\), and \(A_0 = \emptyset\). The historical series \(X_{1:T}\) and the baseline forecast \(F_{\text{base}}\) are immutable components of the workspace; only the editable forecast \(F_t\) and the evidence and trace fields may evolve.

At each step, the agent selects an action
\begin{equation}
    a_t \in \mathcal{A}_{\text{valid}}(W_t) \subseteq \mathcal{A}_{\text{tool}} \cup \mathcal{A}_{\text{revise}} \cup \{\text{stop}\}.
\end{equation}

Tool actions retrieve or summarize contextual evidence and update \(E_t\). Revision actions modify \(F_t\) through a restricted set of forecast-editing operators, such as range-level adjustment or date-specific override. The validity set \(\mathcal{A}_{\text{valid}}(W_t)\) enforces structural constraints: historical observations cannot be changed, the baseline forecast cannot be overwritten, revisions must fall within the forecast horizon, and duplicate or inconsistent edits may be rejected. A valid action induces the transition

\begin{equation}
    W_{t+1} = \mathcal{T}(W_t, a_t).
\end{equation}

The process terminates when the agent selects $\text{stop}$ or reaches a step budget $t^\star$, yielding the output $(F_{\text{final}}, A)$ where $F_{\text{final}} = F_{t^\star}$ and $A = A_{t^\star}$.



Thus, last-mile forecasting is formulated as constrained forecast revision. The agent seeks a final forecast that is better aligned with contextual evidence and operational requirements, subject to validity constraints and accompanied by an explicit trace of the actions that produced it.

\section{Framework}
\subsection{System Overview}


\begin{figure*}[t]  
    \centering
    \includegraphics[width=\textwidth]{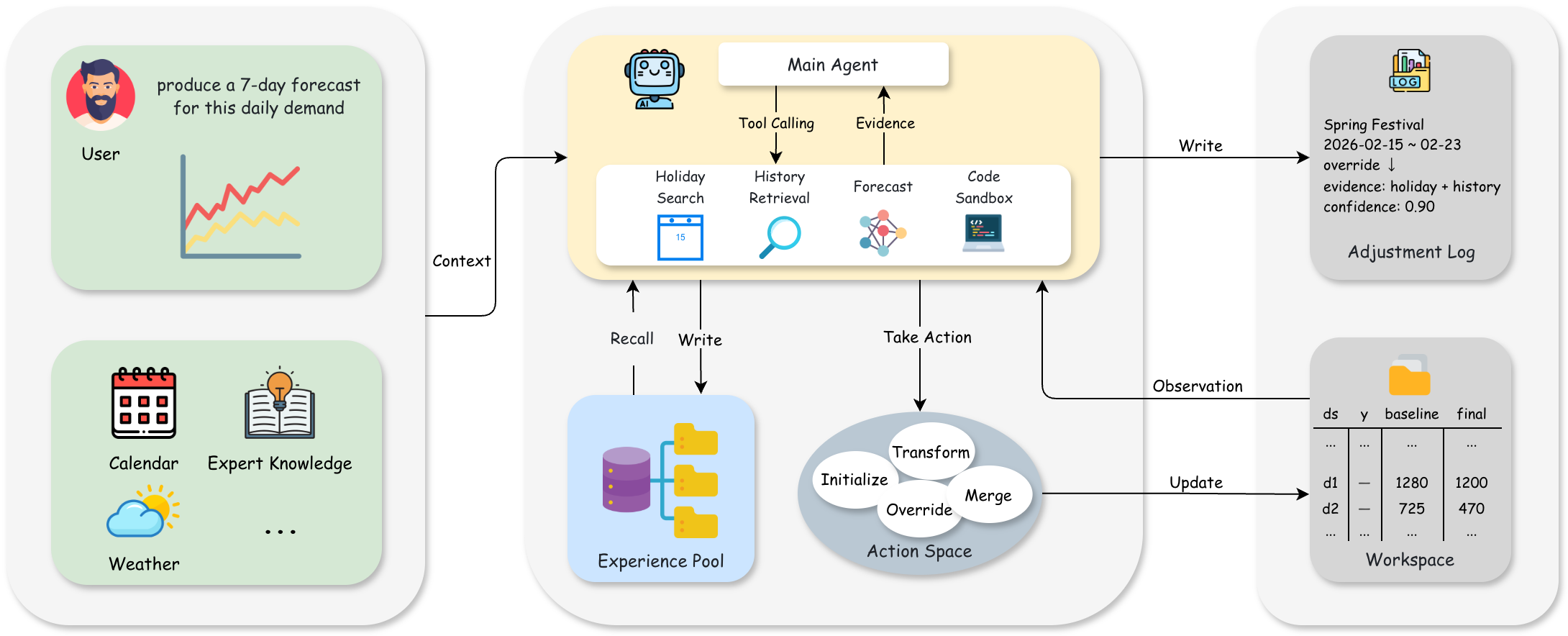}  
    \caption{System overview of the proposed last-mile forecasting framework. A forecasting backbone first produces a baseline trajectory; an LLM agent then operates over a shared forecast workspace, retrieves contextual evidence through tools, applies validated revision actions to \(y_{\text{final}}\), and accumulates reflection memories in a
  persistent memory bank for retrieval by subsequent sessions.}
    \label{fig:system_overview}
\end{figure*}

The framework turns the formulation in Section \ref{sec:problem} into a forecast revision system over a shared workspace. A forecasting backbone first produces a baseline \(F_{\text{base}}\). An LLM-based agent then operates over the workspace state \(W_t\), acquiring contextual evidence, selecting valid revision actions, and updating the editable forecast \(F_t\). Each accepted action is appended to the revision trace \(A_t\), so the final forecast is accompanied by a record of the evidence and operations that produced it.

This design separates numerical extrapolation from contextual revision. The backbone model is responsible for producing a statistically informed baseline, while the agent is responsible for transforming that baseline into a context-consistent and operationally usable forecast. Across forecasting rounds, realized outcomes can also be compared with previous forecasts and stored as reflection memories, which later sessions may retrieve as additional revision evidence.

\subsection{Unified Forecast Workspace}
The central abstraction is a unified forecast workspace. Rather than asking the agent to regenerate a forecast sequence, the workspace represents last-mile forecasting as sequential revision over a shared state. It keeps historical observations, the immutable baseline forecast, and the editable forecast in the same aligned time index, allowing the agent to compare what has been observed, what was originally predicted, and what has already been revised.

Concretely, the workspace is instantiated as a tabular state with four fields:

\begin{table}[htbp]
\centering
\caption{Fields of
  the forecast workspace and their description.}
\begin{tabular}{@{}cc@{}}
\toprule
\textbf{Field} & \textbf{Description}        \\ \midrule
\(ds\)              & timestamp               \\
\(y\)               & historical actual value \\
\(y_{\text{baseline}}\)      & baseline forecast       \\
\(y_{\text{final}}\)         & editable revised forecast \\ \bottomrule
\end{tabular}
\end{table}

Under this representation, \(y\) stores observed series, $y_{\text{baseline}}$ corresponds to the immutable \(F_{\text{base}}\), and \(y_{\text{final}}\) corresponds to the editable forecast \(F_t\). This separation prevents the agent from conflating observations with predictions, preserves a stable baseline for comparison, and localizes all revisions to an explicit output channel. Forecast revision therefore becomes a stateful and auditable process over a shared object rather than an implicit generation process.

\subsection{Constrained Forecast Revision}
The agent is not allowed to replace the forecast with an arbitrary sequence of numbers. Instead, it must revise \(y_{\text{final}}\) through a small action interface. The interface supports four types of operations: initializing \(y_{\text{final}}\) from the baseline, applying a transformation to a date range, replacing selected forecast points, and applying a list of revision actions produced by an earlier decomposition step.

\begin{table}[t]
\centering
\small
\caption{Revision actions supported by the workspace.}
\label{tab:revision-actions}
\begin{tabularx}{\linewidth}{@{}lXX@{}}
\toprule
\textbf{Action} & \textbf{Workspace effect} & \textbf{Use case} \\
\midrule
Initialize & Set \(y_{\text{final}}\) from \(y_{\text{baseline}}\). & Initialize editable forecast objective. \\
Range transform & Apply multiply, add, or clip over a date range. & Holiday, promotion, disruption. \\
Point override & Replace \(y_{\text{final}}\) on selected dates. & Peaks, ramps, specified values. \\
Revisions merge & Apply a list of individual revisions. & Long-horizon forecasting. \\
\bottomrule
\end{tabularx}
\end{table}

Each action specifies where it applies and how it changes the forecast. This makes the revision trace easier to inspect: a range adjustment can be linked to an event window, a point replacement can be linked to a specific date, and a revision list can be traced back to the evidence or planner that produced it. In the notation of Section \ref{sec:problem}, the agent selects an action from \(\mathcal{A}_{\text{valid}}(W_t)\); the workspace is updated only if the action passes validation.

Validation enforces the basic invariants of the workspace. Historical observations cannot be edited, baseline values remain fixed, revisions must lie inside the forecast horizon, and repeated or conflicting edits may be rejected. The agent therefore changes the forecast through checked operations rather than by directly editing the table.

\subsection{Tool-Augmented Evidence Acquisition}
The baseline forecast is produced from the observed series, but many revision decisions depend on information outside that series. The agent therefore uses tools to read different evidence sources, such as past windows in the same workspace, calendar events, external event descriptions, or memories from previous forecast outcomes. Tool outputs are written into the evidence field \(E_t\), rather than directly changing the forecast.

Before evidence can drive a revision, it is converted into a structured revision proposal (Section~\ref{sec:mapreduce} and Appendix~\ref{app:recorded-revisions}). A proposal records the event or context, the affected period, the expected direction of change, an optional magnitude, supporting evidence, and confidence. This structure gives the agent a bridge from contextual information to the revision actions in Table~\ref{tab:revision-actions}: a proposal may justify a range transform, a point override, or a list of decomposed revisions produced by map-reduce planner (Section~\ref{sec:mapreduce}).

\subsection{Long-Horizon Forecasting with Map-Reduce}\label{sec:mapreduce}

\begin{figure*}[t]  
    \centering
    \includegraphics[width=\textwidth]{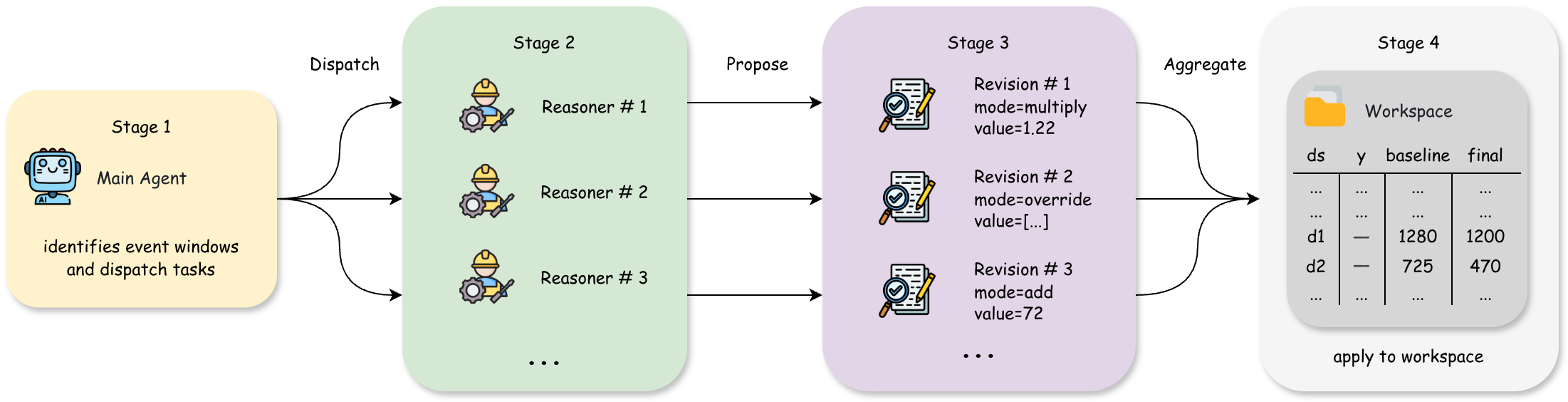}  
    \caption{Map-reduce decomposition for long-horizon forecasting. The
  main agent identifies event windows and dispatches one local
  reasoner per event; reasoners are read-only and emit structured
  revision that are aggregated against the
  workspace through the same constrained action interface used by
  direct revision.}
    \label{fig:map-reduce}
\end{figure*}

Long-horizon revision is harder because relevant events may be spread across many future periods. A single reasoning pass over the full horizon can miss local context or mix unrelated events. To keep the revision process focused, the framework decomposes a long horizon into identified event windows.

For each event window, a local reasoner examines the evidence relevant to that interval and proposes a structured list of revision actions. These event-level plans are then collected and applied to the shared workspace through the same constrained action interface used for direct edits. The decomposition does not introduce a separate way to modify forecasts; it only changes how proposed revisions are generated.

This structure keeps long-horizon revision inspectable. Intermediate plans can be reviewed, filtered, or replayed before they affect \(y_{\text{final}}\), and all accepted revisions must still satisfy the same workspace validity constraints.

\subsection{Memory and Self-Improvement}\label{sec:memory-improve}


The final component of the framework is a post-hoc reflection loop that allows the system to improve its future revisions through memory. Once actual values become available for a previously forecast window, the framework compares the realized outcomes against both the baseline and the revised forecast. This comparison drives a post-hoc reflection step that writes structured experience into a persistent memory bank. The resulting memory capture the event context, the direction and scale of the observed effect, and a concise assessment of how the intervention should have been calibrated when the revision degraded accuracy.

In subsequent forecasting sessions, the agent queries this memory bank before making revision decisions. Retrieved memories serve as empirical priors on event magnitude and revision strategy, complementing the evidence gathered through tools. The framework thereby supports a form of cross-session self-improvement that does not require retraining the forecasting backbone: revision knowledge accumulates from realized outcomes and is reused through retrieval.

Reflection-driven self-improvement has been explored more broadly through experiential-learning frameworks that distill insights from past trajectories \cite{zhao2024expel}. Our setting differs in that experience is grounded in numerically-realized forecast outcomes and stored in a schema-validated form, rather than distilled as free-form natural-language insights from agent trajectories.

\section{Case Studies}

\subsection{Implementation}
We implement the proposed framework with smolagents \cite{smolagents}. The agent is designed as a code‑executing agent equipped with tools for baseline forecasting, historical data retrieval, holiday lookup, memory query, and long‑horizon map‑reduce planning. In this work, TimesFM is used to generate the baseline forecast. The workspace is represented as a dataframe containing the fields \(ds\), \(y\), \(y_{\text{baseline}}\), and \(y_{\text{final}}\), where historical observations and baseline forecasts are treated as immutable. Implementation details, including prompt templates and tool set, are provided in the Appendices \ref{app:setup} and \ref{app:prompts}.

\subsection{Study Design}
\textbf{Case data.} A daily ticket-sales dataset from a popular domestic air route in China, covering 2024-01-01 to 2026-05-05. All three case studies in Sections \ref{study:holiday}–\ref{study:improvement} share this dataset; they differ only in the forecast origin and horizon.

\textbf{Configurations.} For the case studies in Sections \ref{study:holiday} and \ref{study:long}, we evaluate three methods: Prophet \cite{taylor2018forecasting}, TimesFM, and our framework. TimesFM is used without task-specific fine-tuning, while Prophet is fitted on the available history with holiday information. In each case study, all methods are evaluated from the same forecast origin over the same target horizon. Section \ref{study:improvement} reports only our framework, as reflection and memory are internal agent mechanisms rather than numerical forecasting models.

\textbf{Evaluation.} For each comparative case study we report MAE and MAPE on the full forecast horizon and on the event windows defined in the setup. Beyond quantitative error, we discuss the behavior of our framework qualitatively, drawing on the recorded adjustment log and event-level revision records, since auditability and traceability are part of the target behavior.

\subsection{Holiday-Aware Forecast Revision}\label{study:holiday}
This focuses on the Chinese Spring Festival, a major holiday period with strong local demand shifts. All three methods are evaluated from 2026-02-01, over a 23-day horizon ending on 2026-02-23. The horizon covers the days before Spring Festival and the official holiday window from 2026-02-15 to 2026-02-23.

\begin{figure}[t]        
\centering               
\includegraphics[width=1\columnwidth]{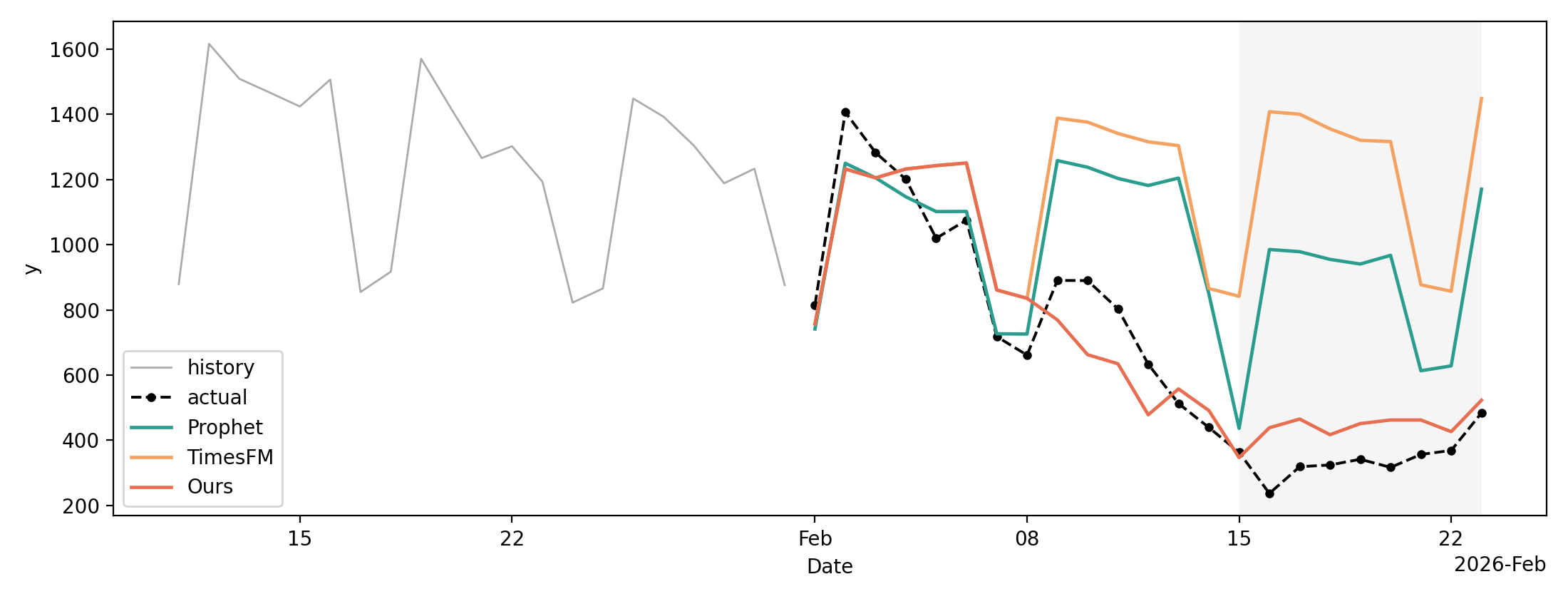}
\caption{Holiday-aware forecast revision}
\label{fig:case_6_2_forecast}
\end{figure}

Figure \ref{fig:case_6_2_forecast} overlays the three forecasts against actual values, with the holiday window shaded; Table \ref{tab:case_6_2_forecast} provides the quantitative comparison. Within the holiday window, our framework reduces MAE by 80.0\% relative to Prophet and by 88.2\% relative to TimesFM, and reduces MAPE from 155.95\% (Prophet) and 262.76\% (TimesFM) to 32.84\%. Over the full horizon, the same ranking holds, demonstrating that the gains in the event window are not paid for by degradation on surrounding non-holiday days.

\begin{table}[htbp]
\centering
{\scriptsize
\setlength{\tabcolsep}{15pt}
\begin{tabular}{@{}lcccc@{}}
\toprule
& \multicolumn{2}{c}{\textbf{Full}} & \multicolumn{2}{c}{\textbf{Holiday}} \\
\cmidrule(lr){2-3} \cmidrule(lr){4-5}
\textbf{Method} & \textbf{MAE} & \textbf{MAPE} & \textbf{MAE} & \textbf{MAPE} \\
\midrule
Prophet & 342.45 & 82.66\% & 507.28 & 155.95\% \\
TimesFM & 530.25 & 131.06\% & 857.28 & 262.76\% \\
Ours    & \textbf{119.17} & \textbf{22.33\%} & \textbf{101.59} & \textbf{32.84\%}\\
\bottomrule
\end{tabular}
}
\caption{Holiday-aware forecast revision}
\label{tab:case_6_2_forecast}
\end{table}

The qualitative behavior is consistent with the recorded revision trace. TimesFM extrapolates the pre-holiday level through the Spring Festival window, while Prophet applies a relatively uniform holiday effect that does not match the depth of this specific holiday trough. Our framework instead applies two override revisions: one for the pre-holiday rundown and one for the official Spring Festival window. Both are grounded in previous-year lunar-calendar analogues and are shown in Appendix \ref{app:spring_festival_revision}. This explains why the revised forecast follows the shape of the realized holiday trough rather than merely shifting the baseline downward.

\subsection{Long-Horizon Event Forecasting}\label{study:long}
This focuses on a longer forecast horizon that contains multiple holiday periods with different local demand patterns. All three methods are evaluated from forecast origin 2025-12-31 over a 126-day horizon ending on 2026-05-05. The horizon contains four event windows: New Year's Day, Spring Festival, Qingming Festival, and Labor Day.

\begin{figure}[t]        
\centering               
\includegraphics[width=1\columnwidth]{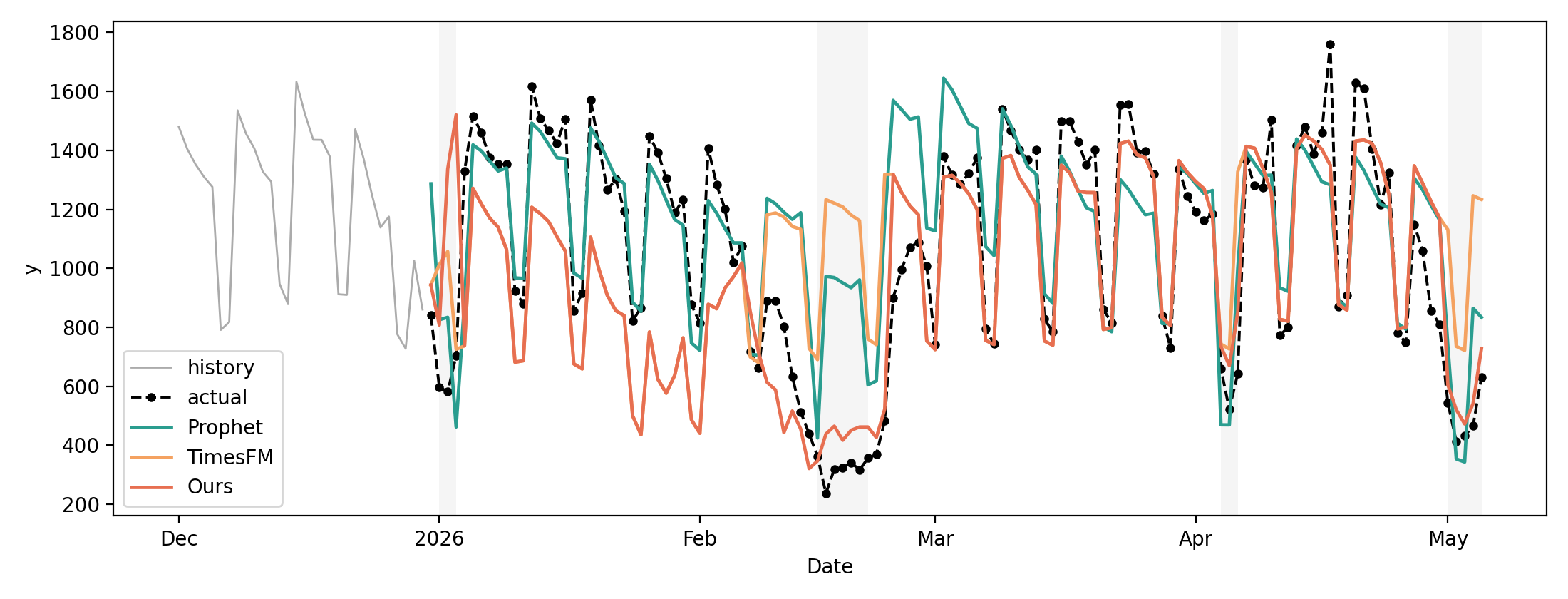}
\caption{Long horizon event forecasting}
\label{fig:case_6_3_forecast}
\end{figure}

We report MAE and MAPE over both the full horizon and each event window. Figure \ref{fig:case_6_3_forecast} overlays the three forecasts across the full horizon, with the event windows shaded; Table \ref{tab:my-table} reports the per-window decomposition. Across the full horizon our framework achieves the lowest MAPE (18.91\%) and a comparable MAE to Prophet (185.7 vs 182.9). Within event windows, it obtains the lowest error on all four holidays, with especially large reductions on New Year, Spring Festival, and Labor Day.


\begin{table}[htbp]
\centering
\label{tab:my-table}
\resizebox{\columnwidth}{!}{%
\begin{tabular}{@{}l*{5}{cc}@{}}
\toprule
& \multicolumn{2}{c}{\textbf{Full}} & \multicolumn{2}{c}{\textbf{New Year}} & \multicolumn{2}{c}{\textbf{Spring Festival}} & \multicolumn{2}{c}{\textbf{Qingming}} & \multicolumn{2}{c}{\textbf{Labor Day}} \\
\cmidrule(lr){2-3} \cmidrule(lr){4-5} \cmidrule(lr){6-7} \cmidrule(lr){8-9} \cmidrule(lr){10-11}
\textbf{Method} & \textbf{MAE} & \textbf{MAPE} & \textbf{MAE} & \textbf{MAPE} & \textbf{MAE} & \textbf{MAPE} & \textbf{MAE} & \textbf{MAPE} & \textbf{MAE} & \textbf{MAPE} \\
\midrule
Prophet & \textbf{182.92} & 27.3\% & 240.65 & 38.7\% & 508.63 & 167.5\% & 199.37 & 31.4\% & 188.31 & 37.6\% \\
TimesFM & 259.81 & 38.1\% & 304.36 & 51.5\% & 742.70 & 241.9\% & 324.18 & 52.8\% & 516.39 & 103.2\% \\
Ours    & 185.72 & \textbf{18.91\%} & \textbf{28.42}  & \textbf{4.65\%} & \textbf{89.28}  & \textbf{29.79\%} & \textbf{169.32} & \textbf{28.68\%} & \textbf{58.96}  & \textbf{12.15\%} \\
\bottomrule
\end{tabular}%
}
\caption{Long horizon event forecasting}
\end{table}

The qualitative evidence supports the intended long-horizon behavior. Rather than applying a single global correction across the 126-day horizon, the framework produces one event-level revision record for each holiday window. As shown in Appendix \ref{app:long_horizon_revision}, the revisions use different operations depending on the local evidence: multiplicative corrections for New Year and Qingming, and per-day overrides for Spring Festival and Labor Day. This event-specific behavior explains why the gains concentrate inside the holiday windows while the full-horizon MAE remains comparable to Prophet.

\subsection{Self-Improvement Mechanism}\label{study:improvement}
This examines whether the framework can learn from a previous adjustment error and use that lesson in a later forecasting turn.

We design a three-week loop on consecutive non-event windows so that the test isolates reflection-driven behavior from event-specific revision. The three windows are W1, W2, and W3; each forecast is produced from an origin one day before the window. After each window's forecast horizon is realized, the framework runs the post-hoc reflection described in Section \ref{sec:memory-improve} and writes a structured entry into the global memory bank. For W2 and W3 we run two configurations: with-memory, which has access to the memory bank populated by previous reflections, and no-memory (control), in which the memory bank is temporarily hidden from the agent.

\begin{table}[htbp]
\centering
{\scriptsize
\setlength{\tabcolsep}{15pt}
\begin{tabular}{@{}lcccc@{}}
\toprule
& \multicolumn{2}{c}{\textbf{no-memory}} & \multicolumn{2}{c}{\textbf{with-memory}} \\
\cmidrule(lr){2-3} \cmidrule(lr){4-5}
\textbf{Window} & \textbf{MAE} & \textbf{MAPE} & \textbf{MAE} & \textbf{MAPE} \\
\midrule
W1 & 33.85 & 2.62\% & -- & -- \\
W2 & \textbf{172.03} & \textbf{12.35\%} & 176.19 & 13.15\% \\
W3 & 92.94 & 7.96\% & \textbf{60.10} & \textbf{5.23\%} \\
\bottomrule
\end{tabular}
}
\caption{Self improvement mechanism}
\label{tab:my-table}
\end{table}

\begin{figure}[t]        
\centering               
\includegraphics[width=1\columnwidth]{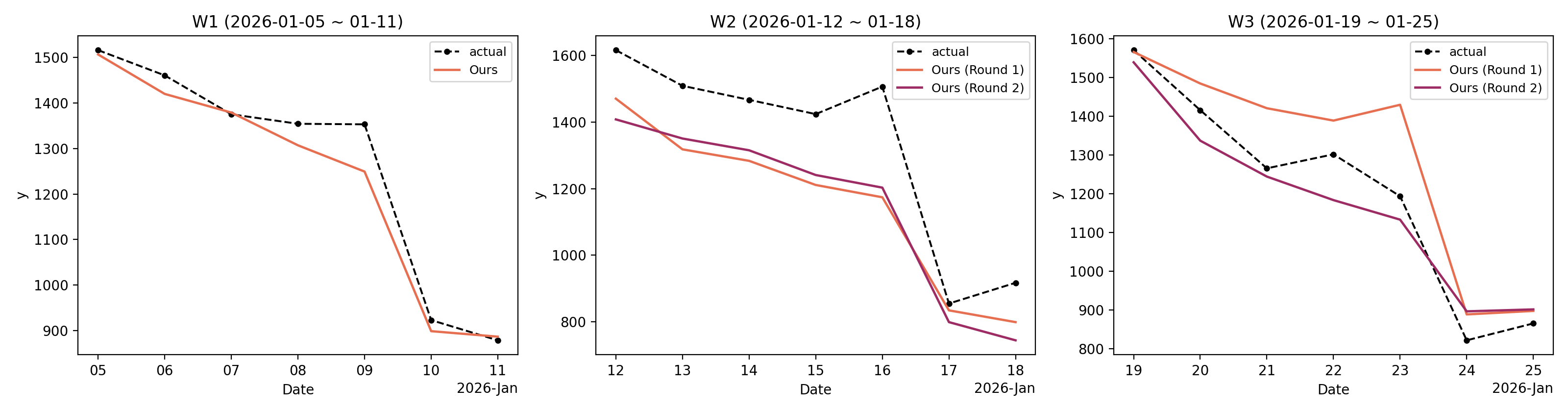}
\caption{Self improvement mechanism}
\label{fig:single_example}
\end{figure}

The memory records provide the qualitative explanation for the W3 improvement. After W1 and W2, the system writes two \texttt{recent\_calibration} entries summarizing realized actual-to-baseline ratios. Appendix \ref{app:reflection_memory} shows that these ratios increase from 1.025 to 1.181, indicating a growing baseline shortfall. The W3 with-memory run can retrieve this experience and use it as a directional prior, whereas the no-memory control only reasons from the current local window.

\section{Conclusion}
We introduced the \textbf{last-mile forecasting} problem and argued that it deserves to be treated as a systematic problem between baseline prediction and business-ready forecasting. To address this problem, we presented an action-centric LLM-agent framework that converts contextual reasoning into constrained, evidence-backed edits over a shared forecast workspace, with support for long-horizon decomposition and cross-session self-improvement.

The core message of this paper is simple: in many practical settings, the hardest part of forecasting is not generating a baseline forecast but turning that forecast into something a human organization can actually use. LLM agents are well suited to this role when they are grounded in structured state, restricted to safe actions, and embedded in an auditable workflow. We hope this perspective encourages future work on forecasting assistants, last-mile benchmarks, and human-agent collaboration for time series decision support.

\section{Limitations and Future Work}
Our current study has several limitations.

First, the paper emphasizes formulation and system design more than large-scale empirical benchmarking. A natural next step is to build a benchmark specifically for last-mile forecasting, where success depends not only on raw numerical accuracy but also on revision correctness, interpretability, and user efficiency.

Second, the current work mainly demonstrates structured textual and calendar-based context. Richer multimodal inputs, such as uploaded planning documents or visual calendar artifacts, remain future work.

Third, external evidence retrieval remains only as reliable as its sources. Future versions should incorporate stronger provenance tracking, confidence calibration, and user approval loops for high-impact revisions.

\nocite{langley00}

\bibliography{example_paper}
\bibliographystyle{icml2026}

\newpage
\appendix
\onecolumn



\section{Experimental Setup Details}\label{app:setup}

\subsection{Dataset}
The case studies use a real-world daily demand series from an industry partner. The full series covers 2024-01-01 to 2026-05-05 (856 values). The series exhibits weekly seasonality, annual seasonality, and clear responses to Chinese-calendar holidays — most visibly the multi-day Spring Festival trough in 2024 and 2025 and the recurring Labor Day and National Day windows.

The dataset has been anonymized before release. Specifically, values are linearly rescaled by an undisclosed positive constant and reported on an arbitrary unit. Calendar dates are preserved so that holiday alignment is reproducible. The relative shape, seasonalities, and event-window responses are preserved; absolute volumes are not interpretable.

\begin{figure}[t]        
\centering               
\includegraphics[width=1\columnwidth]{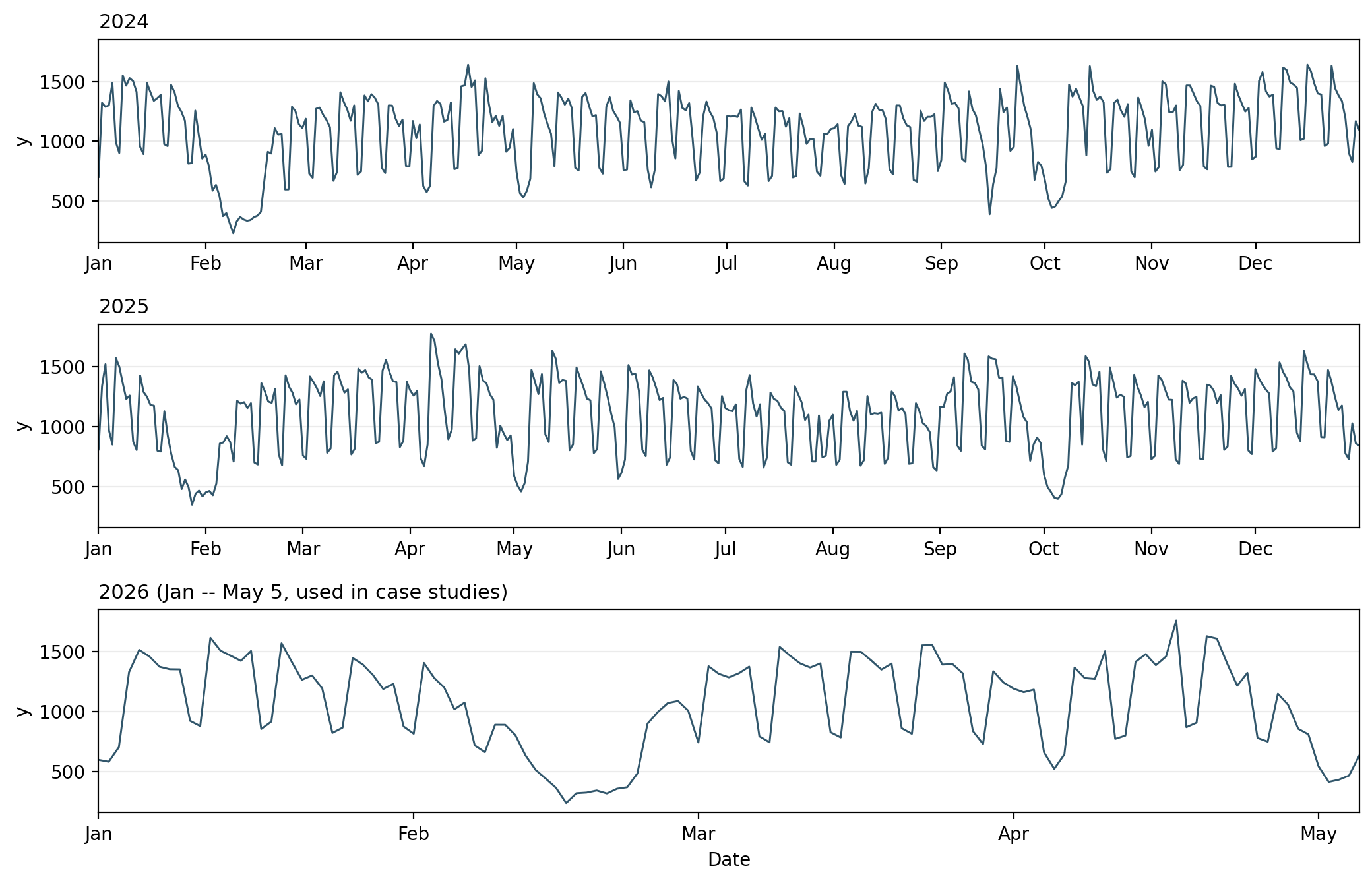}
\caption{Dataset overview}
\label{fig:dataset_overview}
\end{figure}

Figure \ref{fig:dataset_overview} shows the full series. The first two panels cover 2024 and 2025 in their entirety; the third panel covers 2026 up to the latest day used in any case study.

\subsection{Metrics and Event Windows}

We report mean absolute error (MAE) and mean absolute percentage error (MAPE). Metrics are computed over the full forecast horizon and, for event-focused cases, over the event windows listed in Table~\ref{tab:event-windows}. The self-improvement case uses non-event weekly windows, so the full weekly window is also the evaluation window.

\begin{table}[htbp]
\centering
\small
\setlength{\tabcolsep}{19pt}
\begin{tabular}{@{}lp{2.5cm}p{3cm}p{4cm}@{}}
\toprule
\textbf{Case Study} & \textbf{Forecast Origin} & \textbf{Forecast Horizon} & \textbf{Event Windows} \\
\midrule
Holiday-aware revision & 2026-02-01 & 23 days (2026-02-01 -- 02-23) & Spring Festival (2026-02-15 -- 02-23) \\
Long-horizon forecasting & 2025-12-31 & 126 days (2025-12-31 -- 05-05) & New Year (2026-01-01 -- 01-03), Spring Festival (2026-02-15 -- 02-23), Qingming (2026-04-04 -- 04-06), Labor Day (2026-05-01 -- 05-05) \\
Self-improvement (Round 1) & 2026-01-05 & 7 days (2026-01-05 -- 01-11) & -- \\
Self-improvement (Round 2) & 2026-01-12 & 7 days (2026-01-12 -- 01-18) & -- \\
Self-improvement (Round 3) & 2026-01-19 & 7 days (2026-01-19 -- 01-25) & -- \\
\bottomrule
\end{tabular}
\caption{Event Windows}
\label{tab:event-windows}
\end{table}

\section{Agent Prompts and Tool Interface}\label{app:prompts}

\subsection{Main Agent Prompt}
\begin{lstlisting}[caption={Main agent prompt (excerpt).}, label={lst:main-prompt}]
## Role
You are a forecast-revision assistant. A time-series foundation
model provides the numerical baseline; your job is to make evidence-backed,
auditable revisions to that baseline. Do not replace the baseline model.

## Workspace contract
The sandbox exposes a dataframe `df` with columns
  ds, y, y_baseline, y_final
and a helper-owned `adjustment_log`. `last_reflection_summary` carries
post-mortem lessons from previous realized forecasts.

## Non-negotiable rules
- Observed `y` is immutable; existing `y_baseline` is immutable.
- Only revise `y_final`, and only through the provided helpers.
- Every applied revision needs concrete `evidence` and a `confidence`.
- Do not stack multiple revisions on the same range without distinct evidence.
- If the baseline already captures an event, skip the revision.

## Workflow
1. Inspect the series compactly (range, frequency, trend, anomalies).
2. Ensure the baseline exists; otherwise call forecast_tool, then append_forecast.
3. Consult last_reflection_summary; prefer realized lessons over fresh guesses.
4. For each user-mentioned or calendar-relevant event, gather evidence.
5. If the horizon contains MORE THAN ONE event, build a tasks list and call
   run_map_reduce_planners(tasks, context) followed by apply_json_policies.
   For a single isolated event, edit y_final directly via adjust_by_date_range
   or override_forecast_values.
6. Self-review the adjustment_log for empty evidence, implausible impact,
   duplicate ranges, or missing confidence.

## Revision policy (evidence priority)
realized multipliers from reflection > memory critiques > historical
same-period ratios > user instructions.
\end{lstlisting}

\subsection{Local Reasoner Prompt}

\begin{lstlisting}[caption={Local Reasoner prompt (excerpt).}, label={lst:local-reasoner}]
## Role
You are a Local Reasoner dispatched by the orchestrator for one specific
event. Your reasoning ability and evidence access are the same as the
orchestrator's; the differences are:
  1. Your output is a JSON envelope of proposed signals, not direct edits.
  2. Your scope of effect is the assigned event's date range.

## Inputs
The task prompt wraps a global_context block and an assignment block
specifying Event, start_date, end_date (inclusive), and a CSV path holding
the full workspace (history + forecast).

## Reasoning steps
1. Memory first: query_memory_bank(event); a prior critique outranks any
   fresh prior you might pick.
2. Ground the magnitude: retrieve_history_tool over the same-period window
   returned by holiday_search_tool; compute realized/baseline ratio.
3. Check whether the baseline already covers it; if so, propose mode `none`.
4. Pick the shape: range (multiply/add/clip) for uniform effects;
   override (per-day values from a historical analog) for distinctive shapes.
5. Persist the decision via write_signal_envelope(signals); call it once.

## Output schema (per signal)
  source, event, start_date, end_date, mode, value | dates+values,
  direction, magnitude, reason, evidence, confidence
Confidence tiers: 0.9 (two strong evidences agree), 0.7 (one strong),
0.5 (weak/indirect), 0.3 (user instruction only).
\end{lstlisting}
\subsection{Self-improvement Prompt}
\begin{lstlisting}[caption={Self-improvement prompt (excerpt).}, label={lst:self-improvement}]
You are a forecast-revision post-mortem expert. In a previous forecast,
the agent intervened on the following event:
  - Event: {event}
  - Reason: {reason}
  - Action: {code}

Realized comparison (means over the event window):
  - Baseline forecast (y_baseline): {baseline_mean:.2f}
  - Agent-revised (y_final):        {final_mean:.2f}
  - Actual (y_actual):               {actual_mean:.2f}

The agent's intervention increased the error. Summarize the lesson in one
sentence as a critique that should guide future revisions of similar events.
Output a single plain-text sentence, no formatting.
\end{lstlisting}
\subsection{Tool Set}

Table~\ref{tab:tool-set} summarizes the tool interface exposed to the main agent and local reasoners.

\begin{table}[]
\centering
\begin{tabular}{@{}cc@{}}
\toprule
\textbf{Tool}        & \textbf{Role}                                               \\ \midrule
Forecasting tool     & obtains the baseline forecast from the time-series backbone \\
Historical retrieval & retrieves past windows from the current series              \\
Holiday lookup       & provides calendar information for future event windows      \\
Memory query         & retrieves prior reflection entries for similar events and recent experience       \\
Map-reduce planner   & decomposes long horizons into local forecasting intervals      \\ \bottomrule
\end{tabular}
\caption{Tool Set}
\label{tab:tool-set}
\end{table}

\section{Recorded Revisions}\label{app:recorded-revisions}

For each case study we include excerpts from the artifacts produced during the run. The records are condensed for readability: fields not relevant to the discussion are omitted, and free-text reasons are abbreviated.
\subsection{Spring Festival revision}\label{app:spring_festival_revision}

The Section \ref{study:holiday} run produced two accepted revisions, both grounded in the same-period window in the previous Spring Festival.

\begin{lstlisting}[caption={Spring Festival revision: Pre-holiday}, label={lst:case_study_6_2_1}]
event:        Pre-CNY demand rundown
target_range: 2026-02-09 ~ 2026-02-14
mode:         override
reason: Baseline has not captured Spring Festival pre-holiday demand suppression. 2025 YoY analog (Jan 22-27 2025) shows a clear progressive decline from ~769 down to ~478-557, averaging 52% below the baseline.
evidence:
  - holiday_search_tool: 2026-02-15~23 marked pre-CNY workdays;
    same-period reference dates 2025-01-22 ~ 2025-01-27
  - retrieve_history_tool: 2025-01-22~01-27 actuals
    [769.3, 662.4, 634.6, 478.1, 557.4, 491.3]
confidence:   0.8
\end{lstlisting}

\begin{lstlisting}[caption={Spring Festival revision: Holiday}, label={lst:case_study_6_2_2}]
event:        Spring Festival 2026 holiday suppression
target_range: 2026-02-15 ~ 2026-02-23
mode:         override
reason: The baseline (842-1449/day) has NOT captured CNY holiday suppression. 2025 YoY actuals during the corresponding CNY window (Jan 28-Feb 4 2025) ranged from 347 to 523/day (~63% below baseline). Override uses day-by-day 2025 analog values as the 2026 forecast, capturing the characteristic CNY trough shape (deepest on CNY Eve, gradual recovery toward end of holiday week)
evidence:
  - holiday_search_tool: 2026-02-15~23 marked Spring Festival;
    same-period reference dates 2025-01-28 ~ 2025-02-04
  - retrieve_history_tool: 2025-01-28~02-04 actuals
    [346.8, 438.3, 464.9, 417.0, 451.0, 462.1, ...]
  - context: baseline range 842-1449/day during the window;
    holiday suppression not captured
confidence:   0.8
\end{lstlisting}

\subsection{Long-Horizon Revision Records}\label{app:long_horizon_revision}

The Section \ref{study:long} run produced four accepted event-level revision records, one for each holiday window identified by the long-horizon decomposition. We include all four records below.

\begin{lstlisting}[caption={Long-horizon revision: New Year}, label={lst:case_study_6_3_new_year}]
event:        New Year 2026 holiday suppression
target_range: 2026-01-01 ~ 2026-01-02
mode:         multiply (value=0.60)
reason: Jan 1-2 2026 are both New Year holidays. In 2025,
Jan 1 actual demand was 807.6 while Jan 2 normal-workday
demand was 1336.5, giving a holiday ratio of about 0.60.
The 2026 baseline remains near the normal level, so the
record applies a 0.60 multiplier.
evidence:
  - holiday_search_tool: 2026-01-01 and 2026-01-02 are
    both New Year holidays
  - retrieve_history_tool: 2025-01-01 holiday actual=807.6;
    2025-01-02 normal-workday actual=1336.5
  - baseline check: 2026-01-01 baseline=1012.5 and
    2026-01-02 baseline=1057.0, with no clear holiday
    suppression
confidence:   0.7
\end{lstlisting}

\begin{lstlisting}[caption={Long-horizon revision: Spring Festival}, label={lst:case_study_6_3_spring_festival}]
event:        Spring Festival 2026 major demand drop
target_range: 2026-02-15 ~ 2026-02-23
mode:         override
reason: The baseline fails to capture Spring Festival
suppression for most days in the window. The 2025 Spring
Festival analogue averaged 441.1, far below the normal
weekday level. The record uses per-day analogue values,
scaled by the 2026/2025 normal-level ratio, to preserve
the holiday trough shape.
evidence:
  - holiday_search_tool: 2026-02-15~23 are all Spring
    Festival days; same-period reference is 2025-01-28
    ~ 2025-02-04
  - retrieve_history_tool: 2025-01-28~02-04 actuals
    [346.8, 438.3, 464.9, 417.0, 451.0, 462.1, ...],
    avg=441.1
  - baseline check: most 2026 baseline values remain at
    normal weekday levels, so the holiday suppression is
    not captured
confidence:   0.7
\end{lstlisting}

\begin{lstlisting}[caption={Long-horizon revision: Qingming}, label={lst:case_study_6_3_qingming}]
event:        Qingming 2026 over-forecast correction
target_range: 2026-04-06 ~ 2026-04-06
mode:         multiply (value=0.65)
reason: Apr 6 is the last day of the 2026 Qingming holiday.
The baseline value is 1327.4, close to a normal weekday
level, while the same holiday date in 2025 was 849.8,
about 0.65 times the nearby normal weekday average.
evidence:
  - holiday_search_tool: 2026-04-06 is confirmed as a
    Qingming holiday
  - retrieve_history_tool: 2025-04-06 actual=849.8
  - retrieve_history_tool: nearby normal weekday average
    in 2025 is 1307.7, giving ratio 0.650
  - baseline check: 2026-04-06 y_baseline=1327.4, with
    no holiday suppression captured
confidence:   0.7
\end{lstlisting}

\begin{lstlisting}[caption={Long-horizon revision: Labor Day}, label={lst:case_study_6_3_labor_day}]
event:        Labor Day 2026 holiday suppression
target_range: 2026-05-01 ~ 2026-05-05
mode:         override
reason: The 2026 Labor Day baseline significantly
overestimates demand. The same-period 2025 actuals show
a strong holiday suppression pattern, with values between
458 and 706 and an average of 556.1. The record overrides
the forecast with this per-day historical shape.
evidence:
  - holiday_search_tool: 2026-05-01~05 are all Labor Day
    holidays; same-period reference is 2025-05-01~05
  - retrieve_history_tool: 2025-05-01~05 actuals
    [586.4, 504.7, 458.0, 525.1, 706.3], avg=556.1
  - baseline check: 2026 baseline values range from
    721 to 1246, far above the historical suppression
    level
confidence:   0.7
\end{lstlisting}

\subsection{Reflection Memory after the Three-Week Loop}\label{app:reflection_memory}

The Section~\ref{study:improvement} loop writes reflection-memory artifacts after each realized window. Unlike the revision records in the previous subsections, these entries are not direct forecast revisions. They summarize realized calibration evidence that later sessions can retrieve before choosing a new revision strategy. After W2, the memory bank contains two \texttt{recent\_calibration} entries:

\begin{lstlisting}[caption={Reflection memory entries after W2}, label={lst:case_study_6_4_memory}]
entry 1:
  event:       recent_calibration
  range:       2026-01-05 ~ 2026-01-11
  mode:        multiply,  value: 1.025
  evidence:    n_days=7,  mean_baseline=1235.22,
               mean_actual=1265.71,  scale_ratio=1.025

entry 2:
  event:       recent_calibration
  range:       2026-01-12 ~ 2026-01-18
  mode:        multiply,  value: 1.181
  evidence:    n_days=7,  mean_baseline=1123.90,
               mean_actual=1327.35,  scale_ratio=1.181
\end{lstlisting}

These entries are the concrete memory artifacts retrieved by the W3 with-memory agent. The change from 1.025 to 1.181 indicates a growing baseline shortfall, which explains why the later run treats upward calibration as a directional prior rather than relying only on the current local window.

\end{document}